\documentclass[twoside,leqno,twocolumn]{article}
\usepackage{ltexpprt}
\usepackage{subfigure}
\usepackage{graphicx}
\usepackage{amsmath}
\usepackage{amssymb}
\usepackage{bm}
\usepackage{url}
\usepackage{stmaryrd}
\usepackage{balance}
\usepackage{xcolor}

\usepackage{booktabs}
\usepackage{xcolor}
\usepackage{pgffor}
\usepackage{tikz}

\usepackage{epsfig}
\usepackage{pifont}

\newcommand{\ie}[0]{\textit{i.e.}}

\newcommand{\etc}[0]{\textit{etc.}}

\newcommand{\mb}[0]{\mathbf}
\newcommand{\mc}[0]{\mathcal}

\DeclareMathOperator*{\argmax}{arg\,max}

\newcommand{\our}[0]{\textsc{Mpu}}
\newcommand{\well}[0]{\textsc{Well}}

\begin{document}

\title{Large-Scale Multi-Label Learning with Incomplete Label Assignments}
\author{
Xiangnan Kong\thanks{Computer Science Department, University of Illinois at Chicago, USA. xkong4@uic.edu }
\and
Zhaoming Wu \thanks{Computer Science Department, Tsinghua University, China}
\and
Li-Jia Li\thanks{Yahoo! Research, USA. lijiali@yahoo-inc.com}
\and
Ruofei Zhang\thanks{Microsoft, USA.}
\and
Philip S. Yu \thanks{Computer Science Department, University of Illinois at Chicago, USA. psyu@cs.uic.edu}
\and
Hang Wu\thanks{Computer Science Department, Tsinghua University, China}
\and
Wei Fan\thanks{Huawei Noah's Ark Lab, Hong Kong, China. }
}
\date{}
\maketitle

\begin{abstract}
\textit{Multi-label learning} deals with the classification problems where each instance can be assigned with multiple labels simultaneously. 
Conventional multi-label learning approaches mainly focus on exploiting label correlations.
It is usually assumed, explicitly or implicitly, that the label sets for training instances are fully labeled without any missing labels.
However, in many real-world multi-label datasets, the label assignments for training instances can be incomplete.
Some ground-truth labels can be missed by the labeler from the label set.
This problem is especially typical when the number instances is very large, and the labeling cost is very high, 
which makes it almost impossible to get a fully labeled training set.
In this paper, we study the problem of large-scale multi-label learning with incomplete label assignments. 
We propose an approach, called {\our}, based upon positive and unlabeled stochastic gradient descent and stacked models.
Unlike prior works, our method can effectively and efficiently consider missing labels and label correlations simultaneously, and is very scalable, that has linear time complexities over the size of the data.
Extensive experiments on two real-world multi-label datasets show that our {\our} model consistently outperform other commonly-used baselines.
\end{abstract}


\newcommand{\mlknn}[0]{\textsc{Ml-knn}}
\newcommand{\ranksvm}[0]{\textsc{Rank-svm}}
\newcommand{\boos}[0]{\textsc{BoosTexter}}
\newcommand{\cnmf}[0]{\textsc{Cnmf}}
\newcommand{\adt}[0]{\textsc{Adtboost.MH}}
\newcommand{\adaboos}[0]{\textsc{AdaBoost}}
\newcommand{\bsvm}[0]{\textsc{Bsvm}}
\newcommand{\ecc}[0]{\textsc{Ecc}}

\begin{figure}[t]
\centering
    \begin{minipage}[l]{\columnwidth}
      \centering
      \includegraphics[width=\textwidth]{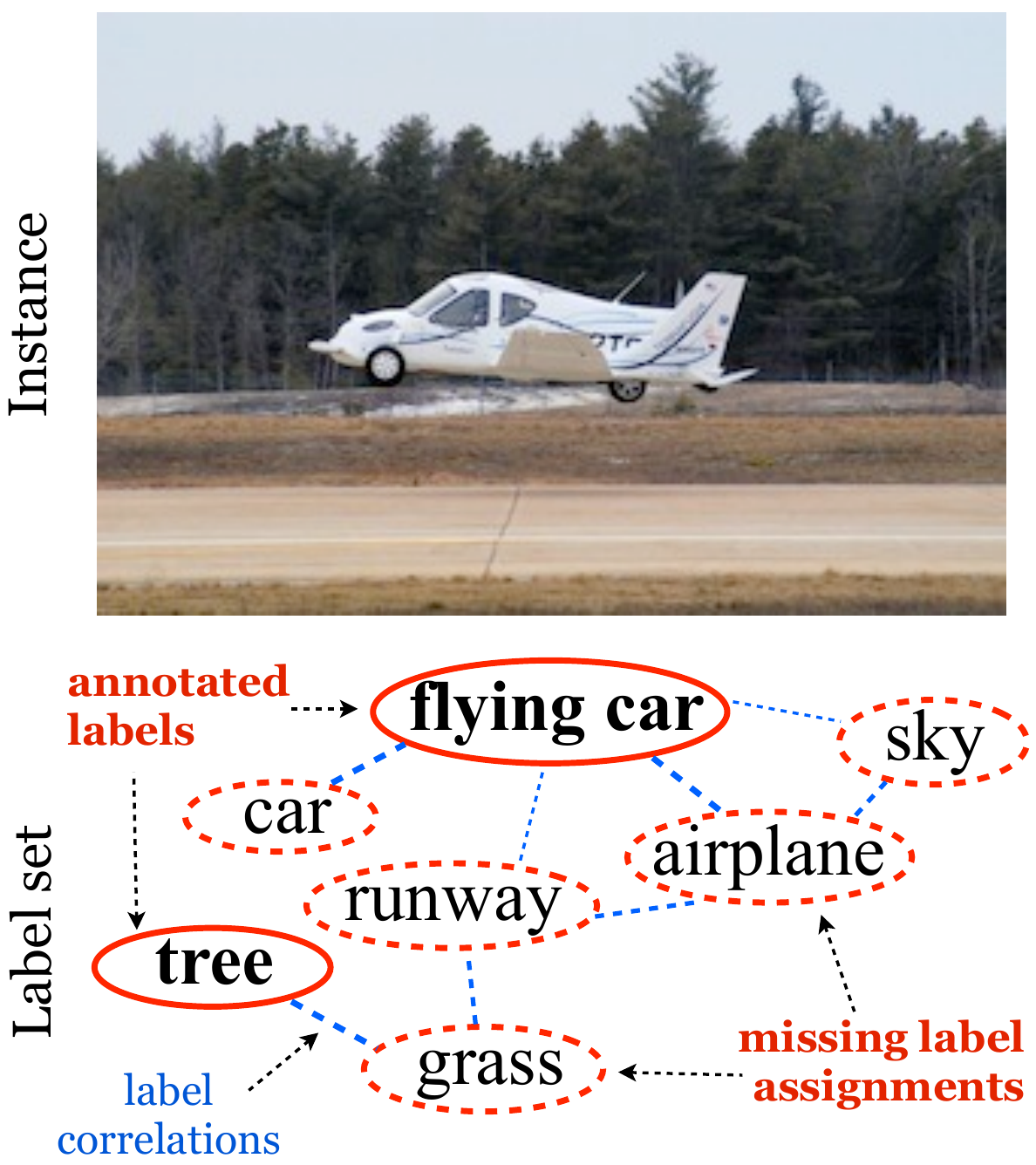}
    \end{minipage}
      \caption{
	      An example of multi-label learning with incomplete label assignments. 
      		The labels annotated by labelers are highlighted by bold font, while the other ground-truth labels are missed by the labelers, but can be inferred by exploiting label correlations.} 
      \label{fig_eg}
\end{figure}

\section{Introduction}
\label{sec:intro}
\textit{Multi-label learning}  has drawn much attention in recent years, where each data example can be assigned with multiple labels simultaneously. 
Multi-label learning has a wide range of real-world applications. 
For example, in text categorization, one text document can belong to multiple categories, such as \emph{sports} and \emph{entertainment};
Researchers in text mining are interested in learning a model that can automatically predict a set of categories for each text document.
Similarly, in image annotation tasks, one image can contain multiple objects or tags, and researchers in computer vision are interested in automatically predicting tags/objects for unlabeled image collections.

Conventional approaches for multi-label learning \cite{US03,GS04,KITM05,ZZ07C,RPHF09} mainly focus on utilizing the correlations among different class labels to facilitate the learning process.
It is usually assumed, explicitly or implicitly, that all the label sets for training instances are fully labeled, {\ie}, all the labels within a label set of an instance should be completely annotated by a labeler without any missing labels. 
However, in many real-world multi-label learning tasks, it is very hard or expensive to get a fully labeled dataset, especially when the number of classes and/or instances are very large.
It is usually much easier to get a set of partially labeled data, where some ground-truth labels for training instances can be missed by the labeler.
For example, in Figure~\ref{fig_eg}, we should an image that contain many tags in its ground-truth label set. 
It usually happens that the labeler may only want to annotate a small number of the labels for the image, instead of going through all possible labels in the vocabulary.
In this example, many true labels of the image are missed by the labeler.
If we directly use existing multi-label learning methods on such datasets, the missing labels in the training data will be treated as negative examples, and the performances of multi-label classification will degenerate greatly due to the simple treatment.

\begin{figure}[t]
\centering
    \begin{minipage}[l]{\columnwidth}
      \centering
      \includegraphics[width=\textwidth]{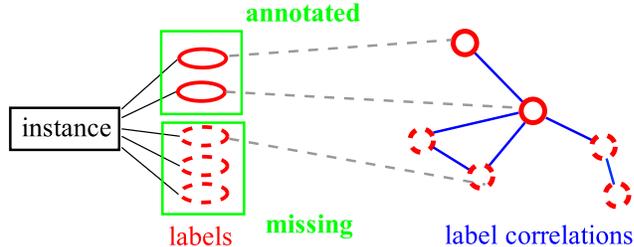}
    \end{minipage}
      \caption{
	      The framework for multi-label learning with incomplete label assignments. } 
      \label{fig_eg2}
\end{figure}

In this paper, we study the problem of large-scale multi-label learning with incomplete label assignments, as shown in Figure~\ref{fig_eg2}.
Despite its value and significance, large-scale multi-label learning  with incomplete label assignments is a much more challenging task due to the specific characteristics of the task. The reasons are listed as follows.

\begin{itemize}
\item \textit{Incomplete Label Assignments}.
Most existing multi-label learning methods, such as {\mlknn} \cite{ZZ07b} and {\ranksvm} \cite{EW02}, assume that the training data are fully labeled.
However, in most real-world applications, the label assignments for training instances can be incomplete.
Thus we cannot simply treat the missing labels as negative examples.
A label that is not annotated by the labeler can still belong the ground-truth label set.
We need to consider the incomplete label assignments explicitly while building our models on training data.
\item \textit{Label Correlations}.
Positive and Unlabeled learning methods \cite{EN08,LDL03} can usually handle the cases when the label assignments are partially missing under single label settings.
However, in multi-label learning problems, each instance can be assigned with multiple labels.
Different class labels are correlated with each other, instead of being independent.
We need to exploit the label correlations to facility the learning process of multi-label classification.
\item \textit{Scalability}.
	Previous approaches on multi-label learning with incomplete label assignments \cite{SZZ10,BJJ11,QYZZ11} are mainly designed for small/moderate-sized datasets.
However, many real-world problems involve a large number of instances. 
In large-scale datasets, it is even more typical to encounter the incomplete labeling issues due to the huge cost of labeling.
In these cases, it is even more important that the learning method can handle large-scale datasets, with time complexities that are linear to both the number of instances and the number of classes.
\end{itemize}

In order to solve the above issues, we propose a novel solution, called {\our}, to learn from partially labeled training instances and can exploit label correlations to facilitate the learning process.
Different from previous work, the proposed {\our} can scale to large-scale problems with time complexity that is linear in both the number instances and number of classes. 
Empirical studies on real-world datasets show that the proposed method can significantly boost the performance of multi-label classification by considering missing labels and incorporating label correlations.

The rest of the paper is organized as follows. 
We start by introducing the preliminary concepts, giving the problem analysis and present the {\our} approach in Section~\ref{sec_model}. 
Then Section~\ref{sec:experiment} reports the experiment results.
We briefly review on related works of multi-label learning and learning from missing labels in Section~\ref{sec:relatedwork}. 
In Section~\ref{sec:conclusion}, we conclude the paper.


\begin{table*}[!ht]
    \centering 
    \caption{Important Notations.}\vspace{10pt}\label{tab:notation}
    \begin{tabular}{|r|l|}
    \hline\vspace{-8pt}&\\
    Symbol& Definition\\
    \hline\vspace{-8pt}&\\
    $\mb{x}_1, \cdots, \mb{x}_n$ & the feature vectors for training instances\\
    $\mb{y}_1, \cdots, \mb{y}_n$   &the set of variables for label sets of the training instances\\
    $\mb{s}_1, \cdots, \mb{s}_n$   &the set of annotated labels for the training instances\\
    $\mb{y}_i=\left(y_i^1, \cdots, y_i^q\right)^\top$ & the vector of variables for the label set of instance $\mb{x}_i$, and $y_i^k\in\{-1,1\}$\\
    $\mb{s}_i =\left(s_i^1, \cdots, s_i^q\right)^\top$ &  the vector of annotated labels for instance $\mb{x}_i$, and $s_i^k \in\{-1,1\}$, $s_i^k \le y_i$\\
    $\mc{D}= \{(\mb{x}_i, \mb{s}_i)\}_{i=1}^n$  &  the  training set for multi-label learning with incomplete label assignments \vspace{2pt}\\
    \hline
    \end{tabular}
\end{table*}

\section{Problem Formulation}
\label{sec_model}

Given $n$ data points of the form $(\mb{x}_i, \mb{y}_i)$, where $\mb{x}_i\in \mc{X}\subseteq\mathbb{R}^D$ is a $D$ dimensional vector denoting the features of the $i$-th instance. 
$\mb{y}_i= (y_i^1, \cdots, y_i^q)^\top \in \{-1, 1\}^q$ corresponds to its set of ground-truth labels within a fixed dictionary of $q$ possible labels. 
$y_i^k=1$ if the $k$-th label is in the label set of instance $i$, otherwise $y_i^k = -1$. 
In many real-world multi-label learning tasks, the training data are usually not fully labeled, where each instance can be labeled with a subset of the ground-truth labels.
We call such settings as multi-label learning with incomplete label assignments or weak labeling problems, following the definition in \cite{SZZ10}.
Specially, in the training set, the ground-truth label set $\mb{y}_i$ for each instance is not available. 
Instead, we only know a set of labels that are annotated by the labeler, denoted as $\mb{s}_i =(s_i^1, \cdots, s_i^q)^\top \in \{-1, 1\}^q$, where
$s_i^k \le y_i^k$ ($\forall 1\le i\le n, \forall 1\le k\le q$).
Thus, $\mb{s}_i$ only provides partial labels for instance $\mb{x}_i$.
When $s_i^k=1$, $y_i^k=1$ is certain.
$$\Pr\!\left( s_i^k = 1 | \mb{x}_i, y_i^k=-1 \right) = 0 ,\ \forall i, k$$
But when $s_i^k=-1$, either $y_i^k=-1$ or $y_i^k=1$ can be true.
In multi-label learning with incomplete label assignments, the training set is a set of partially labeled instances $\mathcal{D} = \left\{ (\mb{x}_i, \mb{s}_i, \mb{y}_i) \right\}_{i=1}^n$, where $\mb{y}_i$'s are unknown.
The learning task is to learn a prediction model $f(\cdot,\cdot)$ from $\mc{D}$, such that for any unseen test data $\mb{x}_i$, the prediction $f(\mb{x}_i, \cdot)$ should be close to the ground-truth, {\ie}, for any $\mb{z}\in\{-1,1\}^q$, $\mb{f}(\mb{x}_i, \mb{z}) = \Pr\!\left(\mb{y}_i=\mb{z} | \mb{x}_i\right)$ as close as possible.

The key issues of multi-label learning with incomplete label assignments are as follows:
\begin{itemize}
\item How can we estimate the true label sets for the training data and use them to facilitate the training process of multi-label learning?
\item How can we exploit the correlations among multiple labels to improve the multi-label classification performances?
\end{itemize}

In the following sections, we will first introduce a model to estimate missing labels in the training set based upon PU ({\ie}, Positive and Unlabeled) stochastic gradient descent.
Next we will describe our framework for incororating the correlations among labels based upon stack models.

\begin{figure*}[t]
\center
\begin{tabular}{l}
\toprule
  \textbf{Input:}\\
  Given a multi-label training set with missing labels $\mc{D}=\{(\mb{x}_i, \mb{s}_i)\}_{i=1}^n$. The number of stacking levels $L$.\\ 
  The base learner $A$, {\ie}, the PU stochastic gradient descent algorithm in Section~\ref{sec_sgd}\\
\textbf{Learning:}\\
\quad - When $\ell=0$, $\forall\ k =1, \cdots, q$, train a model $f_k^{(0)}$:\\  
\quad\quad 1. Construct a training set $\mathcal{D}^{(0)}_k=\left\{(\mathbf{x}_i, s_i^k)\right\}_{i=1}^n$\\
\quad\quad 2. Train a model $f_k^{(0)} = A\!\left(\mc{D}_k^{(0)}\right)$, let $\mb{x}_i^{(0)} = \mb{x}_i$.\\
\quad - Learn the stacked models, for $\ell=1, \cdots, L$ :\\  
\quad\quad 1. Construct estimated predictions $\hat{\bm{y}}_i^{(\ell-1)}$ for $\mc{D}_k^{(\ell-1)}$ using cross-validation in Figure~\ref{alg:cv}\\ 
\quad\quad 2. $\forall\ k =1, \cdots, q$, train a model $f_k^{(\ell)}$:\\
\quad\quad\quad  Construct a extended training set $\mathcal{D}_k^{(\ell)}=\left\{(\mathbf{x}_i^{(\ell)}, s_i^k)\right\}$ \\ 
\quad\quad\quad where $\mathbf{x}_i^{(\ell)} = \left( \mathbf{x}_i^{(\ell-1)}, \hat{\mb{y}}_i^{(\ell-1)}\right)$\\ 
\quad\quad\quad  Let $f_k^{(\ell)} = A\!\left(\mathcal{D}_k^{(\ell)}\right)$ be the  model trained on $\mathcal{D}_k^{(\ell)}$.\\
\textbf{Inference:} given a test instance $\mb{x}$\\
\quad 1. $\hat{\mb{y}}^{(0)} = \left( f^{(0)}_1(\mb{x}), \cdots, f^{(0)}_q (\mb{x}) \right)$, let $\mb{x}^{(0)}=\mb{x}$\\
\quad 2. for $\ell=1, \cdots, L$ :\\
\quad \quad  Construct the extended testing instance $\mb{x}^{(\ell)} =\left( \mathbf{x}^{(\ell-1)}, \hat{\mb{y}}^{(\ell-1)}\right)$ \\
\quad \quad  $\hat{\mb{y}}^{(\ell)} = \left( f^{(\ell)}_1(\mb{x}^{(\ell)}), \cdots, f^{(\ell)}_q (\mb{x}^{(\ell)}) \right)$\\
  \textbf{Output:}\\
  \begin{tabular}{rl}
	  $\hat{\mb{y}}^{(L)}$ :& the label set prediction for the test instance.
    \end{tabular}\\
    \bottomrule
  \end{tabular}
  \caption{The {\our} Algorithm based upon Stacked Graphical Learning and Inference}\label{fig:algorithm}
\end{figure*}


\subsection{Handling Missing Labels via PU Stochastic Gradient Descent}\label{sec_sgd}
In this subsection, we first address the problem of missing label assignments while assuming all labels are independent from each other.
Then in the next subsection, we will show further how to extend the model to consider label correlations.

One simple solution for multi-label learning is to one-vs-all decomposition by treating a multi-label classification problem as multiple binary classification problems (one for each label):
$$
 	\Pr\!\left(\bm{y}_i|\bm{x}_i\right) = \prod\limits_{k=1}^{q} \Pr\!\left(y^k_i=1|\bm{x}_i\right)
$$

Inspired by the positive and unlabeled learning in single-label classification \cite{EN08}, we propose a method, called PU Stochastic Gradient Descent, which can handle large-scale datasets with missing label assignments.
Let $\{\mb{w}_k\}_{k=1}^q$ be a set of parameters of our classifier, where  $\mb{w}_k \in \mathbb{R}^D$.
According to the principle of maximum likelihood, we need to find the optimized $\mb{w}_k^*$ to maximize the likelihood of $y_i^k$'s.
$${\mb{w}_k}^* = \argmax_{\mb{w}_k}\ \log\left( \prod_{i=1}^{n} \Pr\!\left(  y^k_i=1 | \mb{x}_i,\mb{w}_k\right) \right)$$
In this method, we extend logistic regression to classification problems with incomplete label assignments as follows.
Assume that $y_i^k$ satisfies a Bernouli distribution, and 
$$\Pr\!\left( y_i^k = 1 | \mb{x}_i , \mb{w}_k \right) = \frac{1}{ 1 + \exp\!\left( -{\mb{w}_k}^\top \mb{x}_i \right) }$$

Following the assumption in \cite{EN08}, we assume that annotated labels are randomly sampled from the ground-truth label set with a constant rate $c$, where the sampling process is totally independent everything else, such as the feature of the instance.
Assume that the probability that a label is not missing by the labeler is an unknown constant $c$.
$$c = \Pr\!\left( s_i^k = 1 |  y_i^k = 1 \right) = \Pr\!\left( s_i^k = 1 |  y_i^k = 1, \mb{x}_i,\mb{w}_k  \right) $$
Here $c$ can be directly estimated from the training set using cross validation process in \cite{EN08}.
With Bayes' theorem, we have 
$$\Pr\!\left( y_i^k = 1 |  \mb{x}_i,\mb{w}_k \right) = \frac{  \Pr\!\left( s_i^k = 1 |  \mb{x}_i,\mb{w}_k  \right)   }{     \Pr\!\left( s_i^k = 1 |  y_i^k = 1, \mb{x}_i, \mb{w}_k \right) }$$
Then, we have
\begin{align*}
	\Pr\!&\left( s_i^k = 1\ |\ \mb{x}_i,\mb{w}_k \right)  = c \cdot \Pr\!\left( y_i^k = 1 |  \mb{x}_i^k,\mb{w}_k \right) \\
        &=  \frac{c}{    1+\exp\!\left(-{\mb{w}_k}^\top \mb{x}_i \right) }\\
       &=   \frac{c}{    1+\exp\!\left(-s_i^k{\mb{w}_k}^\top \mb{x}_i \right) }  + \frac{(1-s_i^k)(1-c)}{2} \\
	\Pr\!&\left( s_i^k = -1\ |\ \mb{x}_i,\mb{w}_k \right)   = 1- c \cdot \Pr\!\left( y_i^k = 1 |  \mb{x}_i^k,\mb{w}_k \right) \\
       &=   \frac{c}{    1+\exp\!\left(-s_i^k{\mb{w}_k}^\top \mb{x}_i \right) }  + \frac{(1-s_i^k)(1-c)}{2} \\
 \end{align*}
Thus, we can get 
{\small
\begin{align*}
	&{\mb{w}_k}^* = \argmax_{\mb{w}_k} \sum_{i=1}^{n} \log  \Pr\!\left( s_i^k |  \mb{x}_i,\mb{w}_k  \right) \\
			    &= \argmax_{\mb{w}_k} \sum_{i=1}^{n} \log\left(   \frac{c}{    1+\exp\!\left(-{\mb{w}_k}^\top \mb{x}_i \right) } \right)\\
	&=  \argmax_{\mb{w}_k} \sum_{i=1}^{n} \log\left(  \frac{c}{    1+\exp\!\left(-s_i^k{\mb{w}_k}^\top \mb{x}_i \right) }  + \frac{(1-s_i^k)(1-c)}{2}  \right)\\
\end{align*}
}

In order to scale to large-scale problems, we use stochastic gradient descent to solve the above logistic regression problem efficiently.
Different from conventional stochastic gradient descent approaches, which assume all the labels are availabel, we consider the incomplete label assignments and define the loss function as follows:
{\small
$$l(\mb{w}_k, \mc{D}) = -\sum_{i=1}^{n} \log\left(\frac{c}{    1+\exp\!\left(-{s_i^k\mb{w}_k}^\top \mb{x}_i\right) }   + \frac{(1-s_i^k)(1-c)}{2} \right) $$
}

\subsection{ Handling Label Correlations via Stacked Graphical Models}\label{sec:ourmethod}
In the previous subsection, we discussed how can we deal with incomplete label assignments in multi-label learning.
Now we show how can we use the previous model to further consider label correlations.



Inspired by the collective classification methods in \cite{KC07,FJ08} based on stacking, we proposed a multi-label learning method called {\our}. 
{\our} can consider the label correlations effectively using stacked graphical model, which does not rely on joint inference for all labels. 
Stacking \cite{Wolpert92} is one type of ensemble methods which build a chain of models.
Each model in the stacking uses the outputs of previous models as the inputs. 
A stacked multi-label model allows inferences about one label to influence inferences about other labels but uses a different mechanism than other approaches
to multi-label ensemble \cite{DCH10}. 
Rather than using joint inference to propagate inferences among labels, the stacked model uses one base model to predict the class labels for each label and uses those inferred labels as input to another stacked model.

\begin{figure}[t]
\center
\begin{tabular}{l}
\toprule
Given a training set $\mc{D}=\{ (\mb{x}_i, \mb{s}_i )\}_{i=1}^n$ and a PU \\
learning algorithm $A$. Construct a corss-validation\\
prediction $\hat{\mb{y}}_i$ for each instance $\mb{x}_i$ as follows: \\
1. Convert $\mc{D}$ into  $\{\mc{D}_1, \cdots, \mc{D}_q\}$, $\mc{D}_k=\{(\mb{x}_i, s_i^k)\}_{i=1}^n$\\
2. $\forall\ k$, partition $\mc{D}^k$ into $m$ disjoint subsets with \\
\qquad equal-size, denoted as $\mc{D}^1_k, \cdots, \mc{D}^m_k$\\
3. $\forall j=1, \cdots, m, k=1, \cdots, q$, \\
\qquad train one model $f^j_k = A\!\left(\mc{D}_k-\mc{D}^j_k\right)$;\\
\qquad $\forall\ \mb{x}_i \in \mc{D}^j_k$, $\hat{y}_i^k = f^j_k(\mb{x}_i)$; \\
4. $\forall i=1, \cdots, n$, we have $\hat{\mb{y}}_i = \left( \hat{y}_i^1, \cdots, \hat{y}_i^q\right)^\top$\\ 
\qquad Return $\left\{\hat{\mb{y}}_i \right\}_{i=1}^n$\\
\bottomrule
  \end{tabular}
  \caption{Cross-validation to obtain predictions for training instances}\label{alg:cv}
\end{figure}


\noindent\textbf{Learning}:
The learning algorithm to learn the stacked model is shown in Figure~\ref{fig:algorithm}.
First, we use the PU stochestic gradient descent method to learn a base model to predict the labels of the instances using the instance features. 
This base model is then used to infer labels for each of the instances.
In order to avoid overfitting, or any bias from applying the base model on the same data from which it was trained, 
we use a cross-validation procedure (shown in Figure~\ref{alg:cv}) to infer the estimated outputs of the base model as the inputs for the next stacked model.

Next, we then construct an extended feature set to learn a stacked model using both features and estimated outputs of previous models as the inputs. 
In this way, we can build many levels of base models in a stacked model, where each subsequent base model uses the estimated predictions of class labels from the  previous levels.
In most cases, one single level of stacking is sufficient for multi-label learning, that can consider complex label correlations.

\noindent\textbf{Inference}:
During inference process, we take turns to apply the base models from different levels one by one.
The outputs of the model in previous level is directly used as the inputs of the next level in the stacking.
Then the base models in the last level produces the final predictions.
Different from other multi-label ensemble methods that learn on true labels, which are not known at the inference time, 
we learn the stacked model on the inferred labels, where all input features are known at inference time. 
Such design can permit exact inference, while other ensemble methods require approximate inference techniques \cite{DCH10}, such as classifier chains.



\section{Experiments}\label{sec:experiment}

\subsection{Data Collection}

In order to evaluate the performances of the proposed approach for multi-label classification with missing labels, we tested our algorithm on two datasets as summarized in Table~\ref{tab:datastat}.

\begin{itemize}
\item[1)] \textbf{RCV1 Small (Topics Subset)}:
The first dataset we used in this paper is a medium scale dataset, {\ie}, RCV1v2 Topics subset\footnote{\url{http://www.csie.ntu.edu.tw/~cjlin/libsvmtools/datasets/multilabel.html}}, in order to test the performances of different multi-label learning methods on medium scale problems.
The dataset consists of 6,000 news articles which are categorized into 101 classes.
For each news article, the bag-of-words features are extracted resulting in 47,236 features.
\item[2)] \textbf{RCV1 Large (Topics)}: 
The second dataset we used is a large-scale dataset, {\ie}, RCV1v2 Topic full set, in order to test the scalability of different multi-label learning methods on large-scale problems.
The dataset consists of 804,414 news articles which are categorized into 101 classes.
The same number of bag-of-words features are extracted as the previous small dataset.
\end{itemize}

\begin{table}[t]
\caption{Properties of the multi-label datasets.}
\label{tab:datastat}
\centering
\begin{tabular}{lrr}
\toprule
&\multicolumn{2}{c}{Datasets}\\
\cmidrule{2-3}
property &\textbf{RCV1 small}	&\textbf{RCV1 large}	 \\
\midrule 
\# instance	& 6,000 & 804,414 \\
\# label	& 101	& 101 \\
\# feature	& 47,236& 47,236 \\
\bottomrule
\end{tabular}
\end{table}
\begin{table*}[t]
\centering
\caption{Summary of compared methods.}\vspace{10pt}\label{tab:method}
\begin{tabular}{|l|c|l|c|}
\hline&&&\vspace{-8pt}\\
                & \textbf{Type of} 
                & \multicolumn{1}{c|}{\textbf{Properties}}
                & \\
        \textbf{Method}
        &  \textbf{Classification}
        & \multicolumn{1}{c|}{\textbf{Considered}}
        & \textbf{Publication}\\
    \hline\hline \vspace{-10pt}&&&\\
    M3L    & Large-Scale Multi-Label Learning&\ding{172} Label Correlations & \cite{HZVV10}\\ 
	   &&\ding{173} Large-Scale Data &\\
	\hline\vspace{-10pt}&&&\\	
	Elkan08& Positive \& Unlabeled Learning&\ding{174} Missing Label Assignments & \cite{EN08}\\ 
	\hline\vspace{-10pt}&&&\\
	{\well} & Multi-Label Learning with Missing Labels &\ding{172} Label Correlations & \cite{SZZ10}\\
		&&\ding{174} Missing Label Assignments &\\
	\hline\vspace{-10pt}&&&\\
	   &&\ding{172} Label Correlations &\\ 
    {\our} & Large-Scale Multi-Label Learning with Missing Labels&\ding{173} Large-Scale Data & This paper\\
	   &&\ding{174} Missing Label Assignments &\\ 
    \hline
    \end{tabular}
\end{table*}

\subsection{Evaluation Metric}
In order to evaluate the performance of multi-label learning by the models, we follow previous works \cite{GM05,KJS06,LJY06} by using \emph{Micro-F1} as the performance measure.
Suppose that a multi-label dataset is $\mathcal{D}=\left\{ (\mb{x}_i, \mb{y}_i)\right\}_{i=1}^n$, which consists of $n$ multi-label instances.
$\mb{y}_i\in\{0,1\}^q$ ($i=1,\cdots, n$). 
Let $h(\cdot)$  be a multi-label classifier, and its predicted label set for $\mb{x}_i$ is $h(\mb{x}_i)$. 

\emph{Micro-F1} is the harmonic mean of \emph{Micro-Precision} and \emph{Micro-Recall}. The \emph{Micro-Precision} is the Micro-average of precision over all labels
. Similarly, \emph{Micro-Recall} is the Micro-average of recall overal all possible labels.
$$
\text{Micro-F1}(h,\mathcal{D}) = \frac{2\times \sum_{i=1}^{n} \|h(\mb{x}_i) \cap \mb{y}_i \|_1}{\sum_{i=1}^{n} \|h(\mb{x}_i)\|_1 + \sum_{i=1}^{n}\|\mb{y}_i\|_1}
$$ 
The larger the value of \emph{Micro-F1}, the better the performance of multi-label classification model.


All experiments are conducted on machines with Intel Xeon\texttrademark Quad-Core CPUs of 2.26 GHz and 36~GB RAM.

\subsection{Comparative Methods}

In order to study the effectiveness of the proposed approach, we compare our method with different baseline methods, including a large scale multi-label learning method, a multi-label learning method with missing labels and a positive and an unlabeled learning method. 
The compared methods are summarized as follows:
\begin{itemize}
\item\textit{ Large-scale Multi-label Learning (M3L)} \cite{HZVV10}: 
The first baseline method is a  multi-label classification method for large scale problems.
We compared the M3L methods with two different kernels, {\ie}, Linear kernel (denoted as M3L Linear) and RBF kernel (denoted as M3L RBF), separately. 
\item\textit{Positive and Unlabeled Learning (Elkan08)} \cite{EN08}: 
The second baseline method is a PU learning method which handle the cases where some positive instances can be missed by the labeler.
This method is originally designed for binary classification problems.
We use the binary decomposition method to solve multi-label classification problems by training one model over each class \cite{BLSB04}.
\item\textit{Multi-label Learning with Missing Labels ({\well})} \cite{SZZ10}: 
we also compare with another baseline which are designed for multi-label learning with incomplete label assignments.
The method can train a model on weakly labeled multi-label instances, and predict complete label sets on testing data.
It can handle moderate-size datasets, but cannot scale to large-scale datasets.
We use default parameter settings for this method.
\item\textit{Large-scale Multi-label Learning with Missing Labels (\our)}: 
the proposed method in this paper for large-scale multi-label learning with incomplete label assignments. 
{\our} can explicitly consider label correlations to facilitate multi-label learning process when a part of the ground-truth labels are missing in the label set.
For simipicity, we set the number of stacking levels to the minium value 2.
\end{itemize}

\begin{figure}[t]
\centering
    \begin{minipage}[l]{\columnwidth}
      \centering
      \includegraphics[width=\textwidth]{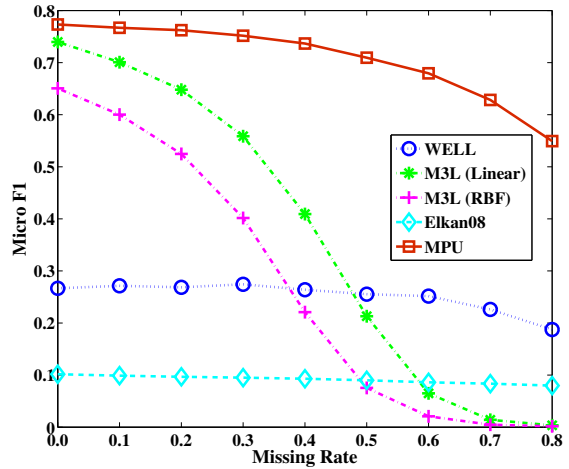}
    \end{minipage}
      \caption{Classification performances on RCV1 small dataset (RCV1 topics subset).} 
      \label{fig_exp_small}
\end{figure}


\subsection{Performances on Small Dataset}
We first study the effectiveness of the proposed {\our} method on multi-label classification with incomplete label assignment. 
In our experiment, $10$-fold cross validation is performed on the small data set to evaluate the multi-label classification performances. 
In each round of the cross validation, the instances are partitioned into two groups: 9 folds are used as training data, the remaining fold is used as testing data.
In order to simulate the incomplete label assignments, we randomly sample and remove a subset of the labels from each of label sets in training data according to a ratio, called \emph{missing rate}. 
For example, if the missing rate is 20\%, we randomly sample 20\% of the labels from the ground-truth labels of training instances and remove them from the train set.
The higher the missing rate, the more ground-truth labels are missed by the labeler in the training set.
We report the average results of 10-fold cross validation on the dataset.

We study the performance of the proposed {\our} method on multi-label classification with different number of missing rates: 0\%, 10\%, 20\%, {\etc}
When the missing rate is 0\%, it represents the setting of conventional multi-label learning where the training instances are fully labeled.
In real-world multi-label learning, we can usually only have a small number of labels annotated in the label set. 

The results of all compared methods are shown in Figure~\ref{fig_exp_small}. 
Firstly, we can observe that when the training set is fully labeled ({\ie}, missing rate is 0\%), all multi-label learning methods outperform the single-label learning method, Elkan08. 
In general, these results support the importance of exploiting correlations among different class labels in multi-label learning problems.
Because the method Elkan08 is based upon one-vs-all decomposition, where different labels are assumed to be independent, thus it cannot work well in multi-label learning tasks when different labels are correlated.

We also observe that when the missing rate increases, the performances of M3L decrease very quickly.
While all the other methods that can consider missing labels are relatively stable in their performances.
This is because M3L is designed for supervised multi-label learning problems, which assumes all the training instances are fully labeled.
When a label $l$ is missed in the label set of a training instance, M3L will consider the instance as a negative example for the label $l$.
This result can support the importance of considering missing labels in the training data, by assuming that some labels in the ground-truth label set can be missed by the labeler.
Thus, for each single class label, the `negative' examples are no longer pure negative examples, but are mixture of both positive examples and negative examples.

We further observe that our proposed method {\our} outperforms all the other methods on all missing rates.
{\our} can estimate the missing rate of the training data automatically, and can utilize label correlations to improve the classification performances.
This result can support the importance of considering both label correlations and missing labels at the same time.

\begin{figure}[t]
\centering
    \begin{minipage}[l]{\columnwidth}
      \centering
      \includegraphics[width=\textwidth]{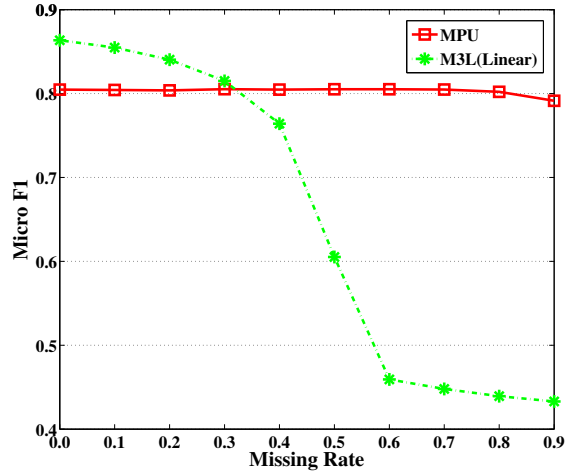}
    \end{minipage}
      \caption{Classification performances on RCV1 large dataset (topics full set).} 
      \label{fig_exp_large}
\end{figure}


\begin{figure}[t]
\centering
    \begin{minipage}[l]{\columnwidth}
      \centering
      \includegraphics[width=\textwidth]{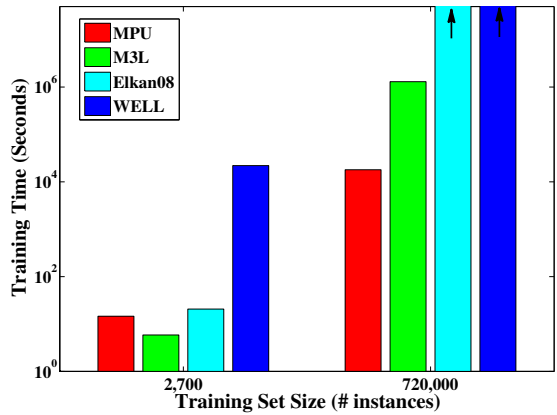}
    \end{minipage}
      \caption{Running time performances.} 
      \label{fig_exp_time}
\end{figure}

\subsection{Performances on Large Dataset}
In our second experiment, we evaluate the efficiency of different multi-label learning methods on a large-scale dataset, which consists of around 800K instances.
In this section, we compare both the Micro-F1 performances and running time in the training step.
We reported the performances of Micro-F1 in Figure~\ref{fig_exp_large}.
We only showed the methods that can finish running within a week.

The methods show similar properties to those in the small datset.
The M3L method initially has good performances when the training set is fully labeled.
However when the missing rate increases, the performances of M3L drop very fast.
The performances of {\our} method is quite stable.
When the missing rate increases, the performances are still pretty good.
{\our} can outperform M3L when the missing rate is greater than 30\%.

We also show the training time of all methods in both dataset in Figure~\ref{fig_exp_time}.
We can observe that both {\our} and M3L can scale well to large-scale datasets.
While the remaining methods such as {\well} cannot scale to datasets in this size.
It is because the time complexity of {\well} is $O(n^2)$ in the number of training instances ($n$). 
Both {\our} and M3L are linear $O(n)$ in the number of instances.

\section{Related Work}
\label{sec:relatedwork}

To the best of our knowledge, this paper is the first work addressing the problem of large-scale multi-label classification with incomplete label assignments. 
Our work is related to both multi-label learning techniques and positive and unlabeled learning. 
We briefly discuss both of them.

Multi-label learning corresponds to the classification problem where each instance can be assigned to multiple labels simultaneously \cite{US03,GS04,KITM05,TV07,GS11}.
The key challenge of multi-label learning is the large space of all label sets, {\ie} the power set of all labels. 
In order to tackle this challenging task, many multi-label learning approaches focus on utilizing the labels correlations to facilitate the learning process. 
Conventional multi-label learning approaches can be roughly categorized as follows:
(a) \textit{one-vs-all} approaches: 
This type of approaches treat  different labels as independent by converting the multi-label problem into multiple binary classification problems (one for each label) \cite{BLSB04}. 
Zhang and Zhou\cite{ZZ07b} proposed {\mlknn}, a binary method by extending the $k$NN algorithm to a multi-label problems using \textit{maximum a posteriori} (MAP) principle to determine the label set predictions. 
(b) \textit{pairwise} correlations: This type of approaches mainly use the pairwise relationships among different labels to facilitate the learning process \cite{GM05}. 
Elisseeff and Weston \cite{EW02} proposed {\ranksvm}, a kernel method by minimizing \textit{ranking loss} to rank label pairs. 
(c) \textit{High-order} correlations \cite{ZZ10,ZZ10}: This type of approaches can utilize  higher order relationships among different labels. 
Examples include random subset ensemble approaches \cite{RPH08,RPHF09}, Bayesian network approach \cite{ZZ10} and full-order approaches \cite{CH09,DCH10}.




The work in \cite{HZVV10} studied the large-scale multi-label leanring problems, and proposed an efficient approach M3L.
In addition, multi-label learning with incomplete label assignment has also been studied on small/moderate-size datasets \cite{SZZ10,BJJ11,QYZZ11}.
The work in \cite{SZZ10} proposed an approach {\well} to infer missing labels in multi-label learning under transductive settings.


Our work is also related to another line of research, called positive and unlabeled learning, or PU learning \cite{LDL03,EN08}.
PU learning corresponds to the binary classification problems where some positive samples can be mislabeled.
Thus in the training set, only positive and unlabeled examples are available, no reliable negative examples are given in the training set.
Many previous works on PU learning focus on estimating reliable negative examples from the unlabeled dataset, and utilize the estimated labels to improve the classification performances,
The work in \cite{LDL03} proposed a method based upon biased SVM. 
Initially all the unlabeled instances are treated as negative examples.
But the cost of classifying an unlabeled example with positive label is lower than that of predicting positive examples with negative labels.
Elkan and Noto \cite{EN08} provided a statistic model for positive and unlabeled learning.
It is assumed that the ground-truth labels are randomly sampled into the training set according to a constant factor.
The random sampling process is assumed to be independent from everything else, such as features of the instances.
The constant factor can be estimated using cross-validation process on the training data.
Then most conventional classification models can be modified according to the factor to consider the missing labels.

\section{Conclusion}\label{sec:conclusion}

In this paper, we have described and studied the problem of large-scale multi-label learning with incomplete label assignments. 
We have studied two real-world datasets, one small and one large to evaluate the performance of our proposed method. 
Different from previous works in multi-label learning, we consider missing labels in the training set and label correlations simultaneously. 
By explicitly consider the missing labels using positive and unlabeled learning model, and label correlations using stacking models, our method can effectively boost the performance of multi-label classification with partially labeled training data.

\section{Acknowledgements}
This work is supported in part by NSF through grants CNS-1115234,  DBI-0960443, and OISE-1129076,  US  Department of Army through grant W911NF-12-1-0066, and Huawei Grant.
\newpage
\balance

\end{document}